\documentclass[11pt]{article}

\usepackage[utf8]{inputenc}
\usepackage[T1]{fontenc}
\usepackage{lmodern}
\usepackage{amsmath}
\usepackage{amssymb}
\usepackage{array}
\usepackage{booktabs}
\usepackage{graphicx}
\usepackage{longtable}
\usepackage[numbers,sort&compress]{natbib}
\usepackage[a4paper,margin=1in]{geometry}
\usepackage{xcolor}
\usepackage{url}
\usepackage{hyperref}

\ifdefined\pdfinfoomitdate
\fi
\ifdefined\pdftrailerid
  \pdftrailerid{}
\fi
\ifdefined\pdfsuppressptexinfo
\fi

\hypersetup{
  colorlinks=true,
  linkcolor=black,
  citecolor=blue!55!black,
  urlcolor=blue!55!black,
  pdfauthor={Xinyi Shan},
  pdftitle={A Pre-Specified Construction-Confirmation Test of Operation-Level Causal Transfer Across Finite Isomorphic Symbolic Domains},
  pdfsubject={Mechanistic interpretability, causal interventions, and independent confirmation}
}

\newcommand{\candidateid}{\texttt{transparent\allowbreak|\allowbreak
integer\_mod16\allowbreak{-}{-}\allowbreak letters16\allowbreak|\allowbreak
successor->predecessor}}
\newcommand{\nnsightstatus}{\texttt{S13\_FULL\_PASS\_SINGLE\_ROUTE}}

\title{A Pre-Specified Construction-Confirmation Test of Operation-Level Causal Transfer Across Finite Isomorphic Symbolic Domains}
\author{
  Xinyi Shan\\
  Independent Researcher\\
  \href{https://orcid.org/0009-0000-6100-1232}{ORCID: 0009-0000-6100-1232}
}
\date{}

\begin{document}
\maketitle

\begin{abstract}
Behavioral accuracy, linear decodability, and successful activation
interventions do not by themselves show that a model carries an operation-level
structure from one symbolic domain to another. We ask a narrower question in
finite isomorphic state spaces: if the hidden-state difference between two
operations is estimated separately for each source input, does adding that
difference to a mapped recipient input move the model toward the corresponding
recipient answer? The design compares this input-specific intervention with
wrong-operation, norm-matched random, and no-op controls, and separates
candidate construction from an independently isolated confirmation split. On a
frozen Qwen2.5-7B-Instruct model at layers 20--21, one route--domain--operation
candidate from a family pre-specified and frozen before confirmation access,
\candidateid, passed both PyVene splits; its confirmation intersection--union
p-value was 0.000198 and its 36-family Holm-adjusted p-value was 0.006943. A
subsequent NNsight 0.7.0 experiment, pre-specified and frozen before its
confirmation access, tested only this selected prompt route, without candidate
or layer reselection. It reproduced all 12 confirmation
effect estimates, confidence intervals, and exact sign-flip p-values
numerically; its 36-family Holm-adjusted p-value was 0.007141. The result is
therefore limited to one prompt route and one candidate, replicated across two
intervention implementations on one model revision and one layer interval. It
does not establish cross-model generalization, full-family backend
independence, domain-general transfer, or algebraic invariance.
\end{abstract}

\section{Introduction}

Interpretability studies often move among questions that require different
evidence. A model may perform a task without representing the task in the way a
researcher expects; a variable may be decodable without being used; and an
intervention may change an answer without transporting a reusable operation.
We focus on the last distinction. The question is not merely whether an
activation patch works, but what is moved, where it is applied, and which
alternative explanations survive the controls.

Consider the operation transfer that ultimately survived the experiment. The
source domain is integers modulo 16 and the recipient domain is the letters
\texttt{a}--\texttt{p}. For source state 3, the ordered transfer
\texttt{successor->predecessor} forms a hidden-state difference between the
predecessor and successor conditions. The same construction is repeated for
every source state rather than averaged into one global direction. The frozen
map sends state 3 to state 8, represented by the recipient symbol \texttt{i}.
Adding the state-3 difference to the recipient representation under the
successor condition asks whether probability moves away from the successor
answer \texttt{j} and toward the predecessor answer \texttt{h}. We call the
per-state difference an \emph{operation contrast}; the domain from which it is
estimated is the \emph{source}, and the mapped domain to which it is applied is
the \emph{recipient}.

This intervention does not copy the activation state of a successful donor run.
It subtracts two operation-conditioned source representations and transports
the resulting difference to a separately mapped recipient input. A \emph{route}
specifies how the operation is worded in the prompt. A formal \emph{candidate}
combines one such route, an ordered source--recipient domain pair, and an ordered
operation transfer. The main effect is compared with a wrong-operation shuffle,
32 norm-matched random directions, and four no-op paths. Construction and
confirmation recompute behavioral eligibility independently and are adjudicated
before their pass sets are intersected.

The narrow contribution is the joint design and its bounded result:
(i) operation contrasts are extracted separately for each source state;
(ii) intervention is recipient-per-input rather than a fixed broadcast;
(iii) operation-specific and no-op controls constrain the same effect;
(iv) construction and confirmation are isolated and independently adjudicated;
(v) exactly one candidate from the pre-specified family passes in both PyVene
splits; and (vi) that selected route passes a separately pre-specified NNsight
construction--confirmation replication without candidate or layer
reselection. None of the ingredients---broad activation steering, successor
representations, isomorphic-domain transfer, controls, or
pre-specification---is claimed as new in itself. The result is one route-bound
candidate, not a universal mechanism.

\begin{figure}[t]
  \centering
  \includegraphics[width=0.96\textwidth]{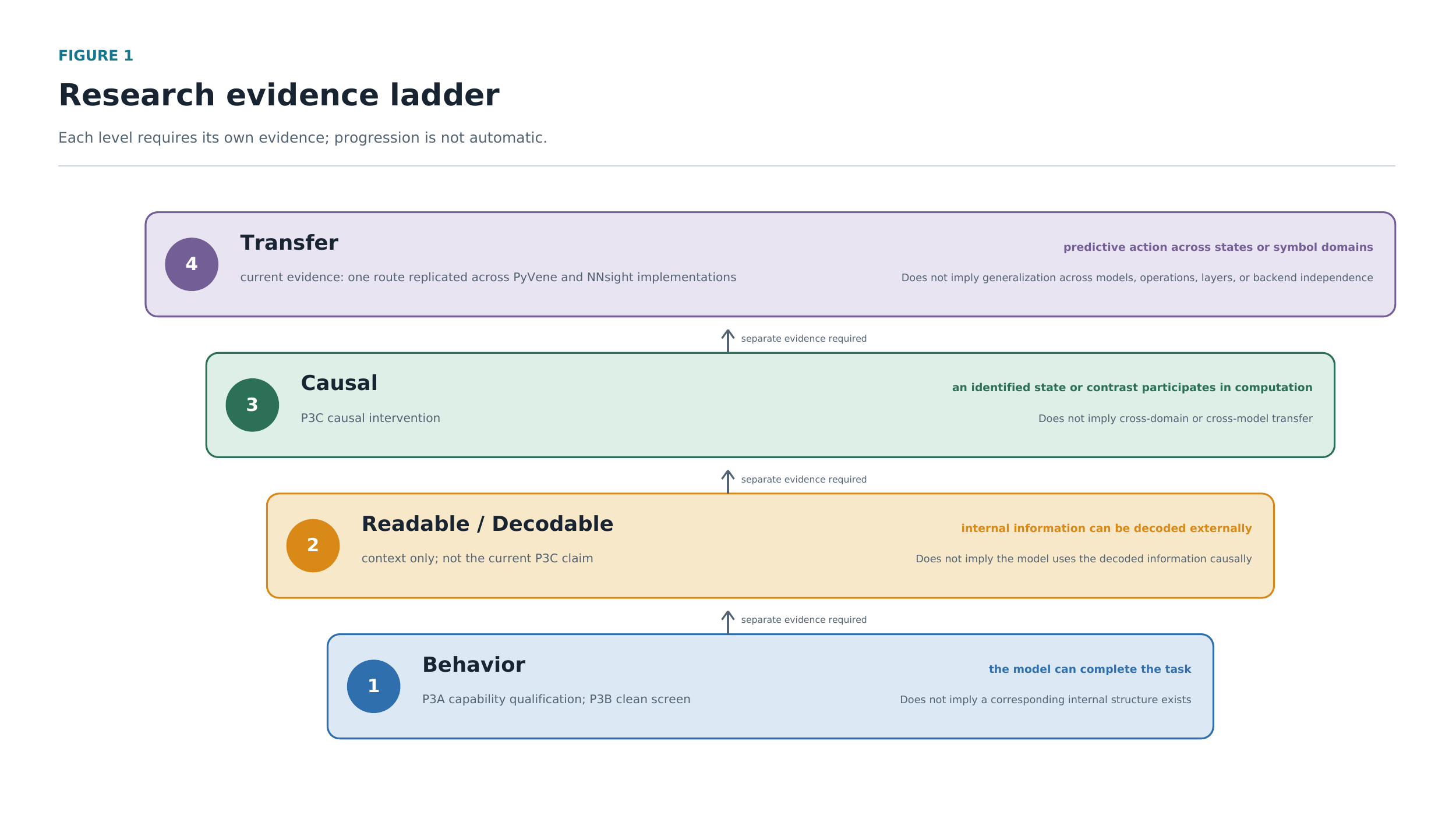}
  \caption{Evidence ladder used by the study. Behavioral success,
  decodability, causal efficacy, and transfer are distinct evidential levels;
  no lower level automatically establishes the next. The current P3C result
  reaches route-bound causal-transfer evidence for one candidate replicated
  across PyVene and NNsight intervention implementations on one frozen model
  revision and layer interval.}
  \label{fig:evidence-ladder}
\end{figure}

\section{Related Work}

\subsection{Function vectors, task vectors, and activation steering}

Several lines of work establish that internal directions can carry tasks or
behaviors and causally affect outputs. Function Vectors use causal mediation
analysis to locate attention heads that
carry task information and combine their outputs into compact vectors that can
induce task behavior under zero-shot and natural-text conditions
\citep{todd2024function}. Task Vectors describe an internal state formed from
demonstrations as a query-independent task representation that can be combined
with a new query \citep{hendel2023taskvectors}. In-context Vectors construct a
latent direction from demonstrations and use it to shift latent states for a
new query \citep{liu2024icv}. Activation engineering constructs inference-time
steering vectors from activation differences between contrastive prompts and
includes random-direction and off-target evaluations
\citep{turner2024activation}. None of this, by itself, distinguishes a fixed
task direction from an input-dependent operation contrast transported through
a prespecified source-recipient route.

\subsection{Successor mechanisms and algorithmic primitives}

Successor Heads identify recurring attention heads that increment ordered
tokens and discuss abstract ordinal and modular features
\citep{gould2024successor}. Emergent Symbolic Mechanisms proposes a staged
architecture of symbol abstraction, induction, and retrieval, and uses causal
patching to analyze successor/predecessor transformations in letter-string
analogies \citep{yang2025symbolic}. Algorithmic Primitives extracts reusable
primitive vectors from reasoning traces and activations and studies
composition by addition, subtraction, and scaling
\citep{lippl2025algorithmic}. The surviving candidate in the present study
contains \texttt{successor->predecessor}, but our experiment does not test
composition, inversion, involution, or preservation of algebraic laws.

\subsection{Causal abstraction and invariance}

Causal abstraction formulates relationships between high-level variables and
neural networks through alignments and interchange interventions rather than
probe accuracy alone \citep{geiger2021causal}. Distributed Alignment Search
extends this framework to high-level variables aligned with distributed neural
representations in a learned basis \citep{geiger2024das}. Causality Is Not
Invariance shows that function-vector interventions can cause behavioral
changes without remaining invariant to input-format changes
\citep{opielka2026causality}. We therefore distinguish intervention efficacy,
split replication, and invariance; the first two do not establish the third.

\subsection{Othello and isomorphic-domain transfer}

Othello World reports a nonlinear representation of board state and uses
interventions to verify its causal role in output behavior
\citep{li2023othello}. mOthello uses token-remapped Othello languages to
separate representational alignment from transfer and reports cross-language
probe interventions \citep{hua2024mothello}. MetaOthello learns an
approximately orthogonal alignment across token-remapped or isomorphic games,
uses residual rotation to predict corresponding actions, and causally steers a
rule branch on held-out ambiguous prefixes \citep{chawla2026metaothello}.
These works are direct precedents for representational alignment and causal
transfer across isomorphic domains; the distinction drawn here is limited to
the joint restriction of a state-specific operation contrast, a
recipient-per-input route, frozen controls, and independently adjudicated
splits.

\subsection{Isomorphic procedures and algebra transport}

Emergent Analogical Reasoning characterizes analogical reasoning using
geometric alignment and functor-like relational transfer
\citep{minegishi2026analogical}. Procedures Across Representations studies
isomorphic procedures and behavioral generalization across code, graphs, and
natural language \citep{lin2026procedures}. Transport of Algebraic Structure
to Latent Embeddings maps input-space operations to latent operations while
requiring specified laws to be preserved \citep{pfrommer2024transport}. These
works prevent ``isomorphic procedures,'' ``functors,'' or ``algebra transport''
from serving as broad contribution language here. The present result tests
neither law preservation nor a formal functor.

\subsection{Cross-format arithmetic mechanisms}

Naganna et al. identify arithmetic heuristic neurons shared across symbolic
arithmetic, natural-language word problems, and Python code in three Llama-3
models. Their causal keep-only and knockout interventions support necessity
and sufficiency within late-layer MLPs. For cross-format transfer, they copy
shared-neuron activation states from a successful donor execution into a
failed target execution, comparing matched same-operator/same-operands donors,
mismatched same-operator/different-operands donors, and random-neuron patches;
matched addition and subtraction recoveries exceed 97\%
\citep{naganna2026form}. This is a direct precedent for shared cross-format
arithmetic mechanisms and causal activation transfer. The distinction retained
here is therefore not cross-format transfer or input matching itself, but the
joint design of transporting a source-derived operation contrast rather than a
donor activation state through explicit finite isomorphisms to mapped recipient
states, together with wrong-operation and no-op controls, isolated construction
and confirmation splits, and frozen family-wise multiplicity correction.

\subsection{Controls, preregistration, and confirmation}

Single-Position Intervention Fails uses zero/random ablation, random sources,
magnitude-matched noise, and shuffled positions
\citep{cheng2026singleposition}. Categorical Perception uses same-norm
random-direction and remapping controls and preregisters a
confirmatory/exploratory hierarchy, bootstrap procedure, and multiplicity
control \citep{cacioli2026categorical}. The Preregistration Revolution
distinguishes using existing observations to generate hypotheses from using
new observations to test predictions \citep{nosek2018preregistration}.
Accordingly, random directions, magnitude matching, shuffling, zero
conditions, causal patching, and preregistration are all direct precedents.
In the reviewed corpus of 20 individually read and archived primary-source
papers, no single paper covered all of these joint restrictions. This is a
corpus-limited observation, not a claim of global novelty, patentability, or
freedom to operate.

\section{Preliminary Design Stages and Pre-Specification}

\subsection{Capability and design gates}

Two preliminary stages separated basic task competence from the later causal
test. The capability-and-measurement stage (P3A) used a smaller 1.5B model to
check that the tasks and measurements were workable. The design-and-screening
stage (P3B) defined the three finite domains and three operations on the 7B
model, then stopped before its planned blinded causal stage because the strict
behavioral screen did not open. Neither stage supplies causal evidence for the
formal causal stage (P3C). Their exact process metrics are retained in
Appendix~\ref{app:preliminary-metrics}; their scientific roles are summarized
below.

\begin{table}[t]
\centering
\small
\caption{Roles of the three stages. Earlier stages do not provide causal
evidence for P3C.}
\label{tab:phase-roles}
\begin{tabular}{llll}
\toprule
Stage & Model & Primary role & Evidential status \\
\midrule
P3A & Qwen2.5-1.5B & Capability and measurement gate & Gate open \\
P3B & Qwen2.5-7B & Domain/operation clean screen & Strict gate closed \\
P3C & Qwen2.5-7B & Causal construction and confirmation & Phase success \\
\bottomrule
\end{tabular}
\end{table}

\subsection{Frozen P3C question}

Before the formal causal data were examined, P3C fixed the candidate universe,
source--recipient mapping, controls, layer-selection rule, behavioral
eligibility rule, construction--confirmation separation, and 36-candidate Holm
procedure. The diagnostic route was also declared ineligible for promotion.
These choices prevent candidate selection and success criteria from being
adapted to the confirmation outcome.

\section{Methods}

\subsection{State spaces, operations, routes, and candidates}

The study used three 16-state domains sharing abstract state
$x\in\{0,\ldots,15\}$:
\texttt{binary4} (four-bit strings \texttt{0000}--\texttt{1111}),
\texttt{integer\_mod16} (integers 0--15), and
\texttt{letters16} (letters a--p, with a$=0,\ldots,$p$=15$). The frozen
operations were
\[
\operatorname{successor}(x)=(x+1)\bmod16,\quad
\operatorname{predecessor}(x)=(x-1)\bmod16,\quad
\operatorname{reflection}(x)=15-x.
\]

Formal candidates were evaluated under two prompt-wording routes pre-specified
and frozen before confirmation access. The \texttt{transparent} route specified the complete surface recipe,
domain, wrap behavior, and output form. The \texttt{relational} route specified
the complete next/previous/mirror relation without requiring abstract-index
calculation. A third \texttt{abstract} route used an explicit
$\mathbb{Z}_{16}$ formula with surface encoding and decoding; it was diagnostic
only and could not promote a candidate.

The formal family contained 36 candidates: two formal routes, three unordered
domain pairs, and six ordered operation transfers. Within a domain pair,
\texttt{D1} and \texttt{D2} denote the two frozen domain labels; $A$ and $B$
denote the ordered source and target operations. Behavioral eligibility was
computed independently in construction and confirmation. For each candidate,
all four domain-by-operation slices (\texttt{D1xA}, \texttt{D1xB},
\texttt{D2xA}, and \texttt{D2xB}) had to achieve at least 14/16 canonical
exact. Each measurable state additionally required the necessary clean samples
to be canonical exact, distinct recipient outputs, and a finite oriented
denominator greater than $10^{-6}$.

All 16 states of \texttt{successor<->predecessor} were identifiable, and at
least 12/16 states had to be eligible in both directions. Each other unordered
operation pair excluded two collision states and required at least 12/14
eligible states in both directions. Only eligible candidates proceeded to
causal measurement. Ineligible or unmeasured candidates remained in the
36-candidate statistical family with $p=1$.

\subsection{Input-dependent operation contrasts}

For source domain $S$, layer $l$, ordered transfer $A\rightarrow B$, and
state $x$, define
\[
\Delta_{S,l}(A\rightarrow B,x)
=h_{S,l}(B,x)-h_{S,l}(A,x).
\]
The frozen cross-state map was
\[
\pi(x)=(x+5)\bmod16.
\]
The contrast was added to the same-layer representation of recipient domain
$R$ at the distinct state $\pi(x)$:
\[
h^{\mathrm{patched}}_{R,l}(A,\pi(x))
=h_{R,l}(A,\pi(x))+\Delta_{S,l}(A\rightarrow B,x).
\]
Thus, each source input supplied its own operation difference; the method did
not reuse a donor's full activation state or broadcast one direction to all
recipient inputs. The patch representation was the transformer-block output.
The patch site was the single residual-stream vector at the
\texttt{assistant\_output\_boundary}; no pooling was used. The two
source--recipient directions were measured separately.

\begin{figure}[t]
  \centering
  \includegraphics[width=0.97\textwidth]{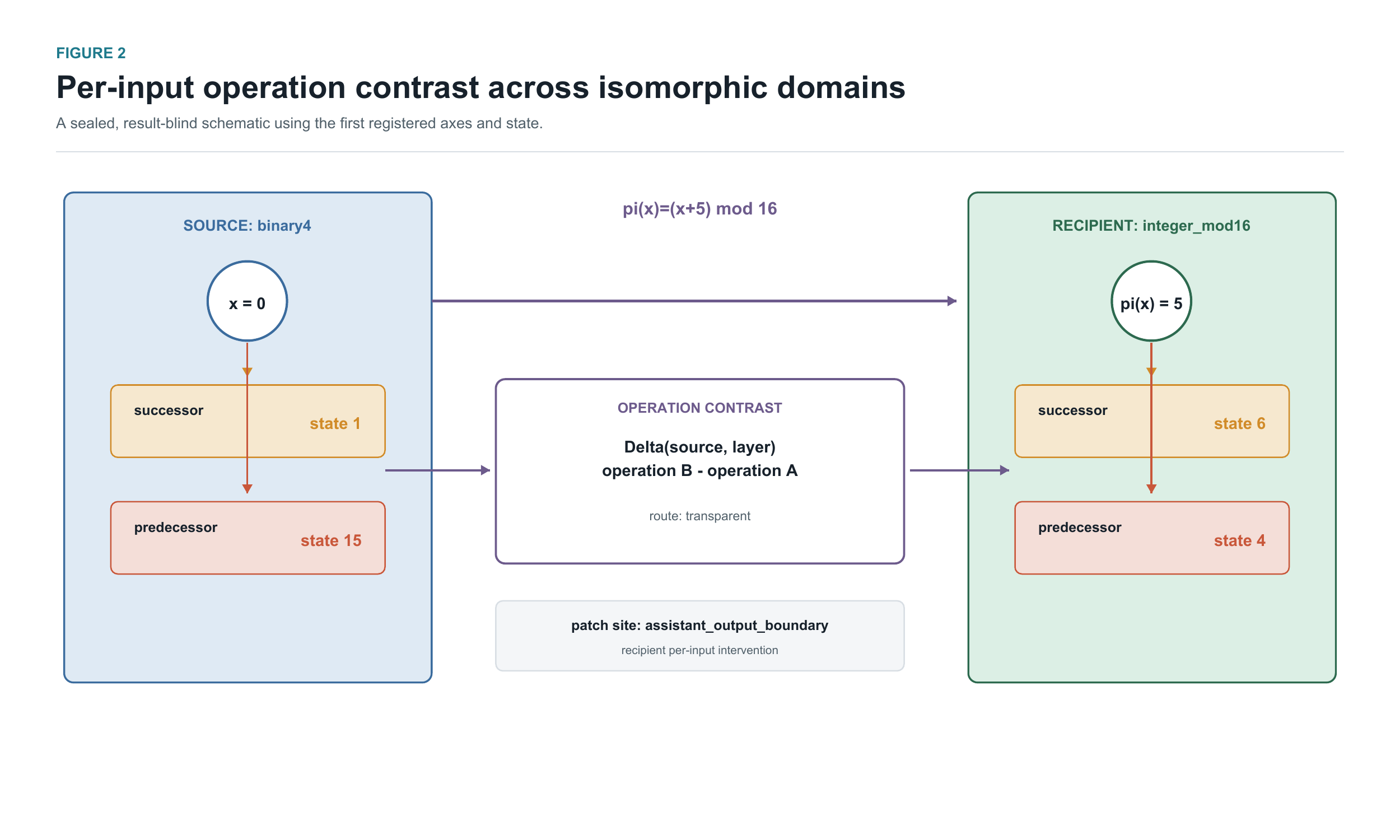}
  \caption{Result-blind schematic of the recipient-per-input operation
  contrast. A source-domain difference between operations $B$ and $A$ is
  extracted separately for each state $x$, then added to the recipient-domain
  representation at $\pi(x)=(x+5)\bmod16$. The schematic does not select a
  route or candidate from the observed results.}
  \label{fig:operation-contrast}
\end{figure}

\subsection{Control interventions and norm matching}

Each measured state--direction--layer bucket contained exactly 38 records:
one real contrast, one shuffled-operation contrast, 32 deterministic
matched-random contrasts, and four no-op paths (zero, identity, inline
self-replacement, and cached self-replacement). Matched-random used seed 619
and replicate identifiers 0--31. Each seed preimage bound the sample, domains,
operations, layer, patch site, and replicate.

The shuffled control replaced the requested operation transfer with a
different ordered transfer. The matched-random controls tested whether a
direction of the same magnitude could produce the effect without carrying the
source operation contrast. Zero, identity, and the two self-replacement paths
tested whether the patching pipeline itself changed the output.

For matched-random and shuffled-operation controls, norm matching followed an
FP16 rule pre-specified and frozen before confirmation access. The L2 norm of each FP16 tensor was accumulated in
float32 and then represented as an FP16 scalar. A control was accepted only
when this scalar had exactly the same FP16 bit pattern as the corresponding
real contrast; otherwise the bucket was rejected rather than analyzed.

\subsection{Model, backends, generation, and layer selection}

Formal P3C used \nolinkurl{Qwen/Qwen2.5-7B-Instruct} at revision
\texttt{a09a35458c70\allowbreak{}2b33eeacc393\allowbreak{}d103063234e8\allowbreak{}bc28}.
The model files matched a 14-member per-file manifest. Full-family construction
and confirmation used PyVene 0.1.8 at commit
\texttt{8138d93a7e5b\allowbreak{}1b5742d71f9f\allowbreak{}750447f2e1b6\allowbreak{}5b95}.
The environment used Python 3.11.9, PyTorch 2.11.0+cu128, Transformers 4.57.6,
Tokenizers 0.22.2, an NVIDIA GeForce RTX 3090, and NVIDIA driver 580.105.08.
The PyTorch build targeted CUDA 12.8; the driver reported CUDA compatibility
13.0.

The subsequent selected-route replication used NNsight 0.7.0 with the same
model revision and layers. Its environment used Python 3.11.9, PyTorch
2.11.0+cu128, Transformers 4.57.6, Tokenizers 0.22.2, and an NVIDIA GeForce RTX
3090. The environment inventory, model files, package, and source-tree
identities were recorded before scientific execution.

Deterministic algorithms and \texttt{CUBLAS\_WORKSPACE\_CONFIG=:4096:8}
were enabled. CUDA matmul TF32, cuDNN TF32, Flash SDP, and memory-efficient SDP
were disabled; math SDP was enabled. Clean generation used greedy token-wise
argmax for at most 16 new tokens and stopped at EOS or the first newline in the
accumulated decoded text. No answer normalization was applied. Sequence scores
summed token log probabilities for the complete candidate answer followed
directly by EOS.

PyVene construction localized one global interval across all 28 transformer
blocks and selected zero-based layers $[20,21]$. The number of blocks is an
execution fact for this model, not a cross-model protocol constant. PyVene
confirmation and the NNsight replication both reused this interval without
layer reselection.
\subsection{Construction and independent confirmation}

Construction completed its clean battery and split-specific eligibility before
performing one global localization. Operation-recovery support required a
layer mean of at least 0.70 of the global peak and a clustered-bootstrap lower
95\% bound strictly greater than zero. Answer-transfer onset was the earliest
position at which two consecutive blocks reached 0.50 of the global peak.

A candidate interval had to end before answer-transfer onset, be contiguous,
and have width at least two. Ties were broken by highest mean operation
recovery, earliest start, and shortest width, in that order. The entire
candidate universe shared one global interval; candidate-specific layer
selection was prohibited. Construction selected layers $[20,21]$ and observed
answer-transfer onset at layer 22.

Confirmation inputs were prepared and kept inaccessible before construction
results existed. They became available to the analysis only after the
construction pass set and layers had been fixed. Confirmation then recomputed
clean eligibility, measured eligible formal candidates, and applied the same
statistical family. Diagnostic results remained separate and could not promote
a candidate. This split isolation constrains adaptation to confirmation data; it
is a validation property, not part of the claimed neural mechanism.

\subsection{Effects and statistical adjudication}

For target operation $B$ relative to source operation $A$, define the
full-answer sequence margin
\[
M(\mathrm{run};B,A)=
\log P\!\left(y_{R,B}(\pi(x))\,\mathrm{EOS}\mid\mathrm{run}\right)
-\log P\!\left(y_{R,A}(\pi(x))\,\mathrm{EOS}\mid\mathrm{run}\right).
\]
The raw margin shift and normalized recovery were
\[
dM=M(\mathrm{patched};B,A)-M(\mathrm{clean}_A;B,A),
\]
\[
R=\frac{dM}
{M(\mathrm{clean}_B;B,A)-M(\mathrm{clean}_A;B,A)}.
\]
The denominator had to be finite and strictly greater than $10^{-6}$;
recovery was not clipped. Layers and eligible states were equally weighted.

Four effect quantities were computed: real $dM$, real normalized recovery,
real-minus-mean-matched-random recovery, and real-minus-shuffled-operation
recovery. Each quantity was summarized for \texttt{D1\_to\_D2},
\texttt{D2\_to\_D1}, and their bidirectional aggregate. We call each such
effect--direction summary a \emph{component}, giving 12 pre-specified components
per candidate.

Each component used a 10,000-resample state-clustered bootstrap for a 95\%
confidence interval. The cluster unit was the abstract-state bidirectional
atomic pair. The exact one-sided state-level sign-flip test enumerated all
$2^n$ sign assignments and computed
$P(T_{\mathrm{flip}}\ge T_{\mathrm{observed}})$. At the candidate level,
the intersection--union test (IUT) requires every component to pass, so its
p-value is the largest component p-value:
\[
p_{\mathrm{IUT}}=\max_{k=1,\ldots,12}p_k.
\]
The \emph{Holm family} is the full set of 36 formal candidates. Holm step-down
correction controlled family-wise error across that set. A candidate passed
only if all 12 lower confidence bounds were strictly positive, all four no-op
checks passed, the Holm-adjusted p-value was at most 0.05, and the candidate was
measured.

\subsection{NNsight selected-route replication protocol}

After the PyVene construction--confirmation intersection was frozen, a
separate chain, pre-specified and frozen before NNsight confirmation access,
tested only \candidateid\ with NNsight 0.7.0. It
used the same model revision, layers $[20,21]$, source/recipient directions,
state roster, causal estimands, matched-random and shuffled-operation controls,
four no-op paths, clustered bootstrap rule, and exact sign-flip test. Each
split contained $16$ states $\times 2$ directions $\times 2$ layers $\times
38$ records = 2,432 causal/control rows. The protocol prohibited candidate or
layer reselection and prohibited restarting the scientific chain after it had
begun.

This was not a second 36-candidate discovery or screening pass. The frozen
36-member multiplicity family was retained: the selected candidate received
its measured IUT p-value, while each of the other 35 unmeasured family members
was assigned $p=1$. The same Holm implementation was then applied to all 36
members.

\section{Results}

\subsection{Construction and confirmation flow}

Construction yielded 12 eligible candidates and five passes. Confirmation
contained 432 unique clean-prediction records:
$3$ routes $\times 3$ domains $\times 3$ operations $\times 16$ states.
Formal exact accuracy was 243/288, diagnostic exact accuracy was 95/144, and
overall exact accuracy was 338/432. Six formal candidates were eligible in
confirmation, and three passed Holm correction. The pass sets were computed
independently and intersected only after split-specific adjudication:
\[
|\mathcal{P}_{\mathrm{construction}}|=5,\quad
|\mathcal{P}_{\mathrm{confirmation}}|=3,\quad
|\mathcal{P}_{\mathrm{construction}}\cap
\mathcal{P}_{\mathrm{confirmation}}|=1.
\]

\begin{figure}[t]
  \centering
  \includegraphics[width=0.98\textwidth]{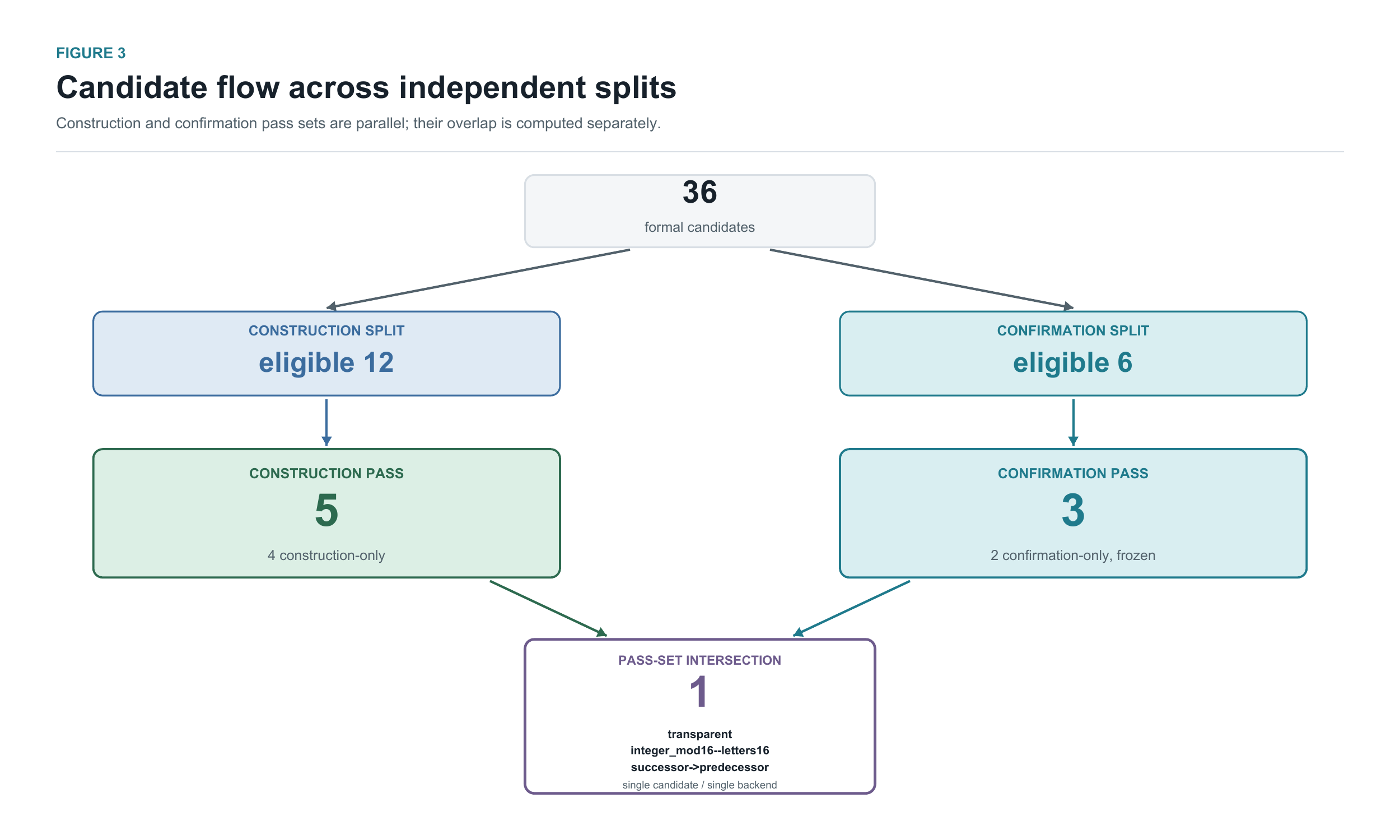}
  \caption{PyVene full-family construction and confirmation flow. Eligibility is
  recomputed in each split: $36\rightarrow12\rightarrow5$ in construction and
  $36\rightarrow6\rightarrow3$ in confirmation. Only the pass sets are then
  intersected, yielding one both-split candidate. Four candidates are
  construction-only; two confirmation-only positives remain frozen,
  uninterpreted, and unpromoted. No funnel, inheritance, or subset relation
  between split-specific eligibility sets is implied.}
  \label{fig:split-flow}
\end{figure}

The unique intersecting candidate was \candidateid. The two
confirmation-only passes were
\texttt{relational\allowbreak|\allowbreak binary4\allowbreak{-}{-}\allowbreak
integer\_mod16\allowbreak|\allowbreak successor->reflection} and
\texttt{relational\allowbreak|\allowbreak
integer\_mod16\allowbreak{-}{-}\allowbreak letters16\allowbreak|\allowbreak
successor->predecessor}.
They remain frozen, uninterpreted, and unpromoted.

\subsection{Causal and control records}

The confirmation causal/control dataset contained 12,768 records:
336 real, 336 shuffled-operation, 10,752 matched-random, and 336 records for
each of the four no-op classes. The 1,344 no-op effects were all exactly zero.

\begin{table}[t]
\centering
\small
\caption{Confirmation causal and control records.}
\label{tab:record-composition}
\begin{tabular}{lr}
\toprule
Record class & Count \\
\midrule
Real & 336 \\
Shuffled operation & 336 \\
Matched random & 10,752 \\
Zero delta & 336 \\
Identity patch & 336 \\
Inline self-replacement & 336 \\
Cached self-replacement & 336 \\
\midrule
Total & 12,768 \\
\bottomrule
\end{tabular}
\end{table}

\subsection{The PyVene split-replicated candidate}

For the candidate
\begin{center}
\small\candidateid
\end{center}
the intersection--union p-value was 0.0001983642578125 and the 36-family
Holm-adjusted
p-value was 0.0069427490234375. Aggregate normalized recovery was 0.55358
[0.45874, 0.65043], real-minus-matched-random recovery was 0.51961
[0.42152, 0.62057], real-minus-shuffled recovery was 0.26683
[0.19295, 0.33861], and real $dM$ was 17.74475
[14.17413, 21.17219]. The weakest constrained directional component was
\texttt{D2\_to\_D1} real-minus-shuffled recovery: 0.15868
[0.09283, 0.23190], with exact $p=0.0001983642578125$.

\begin{figure}[t]
  \centering
  \includegraphics[width=0.98\textwidth]{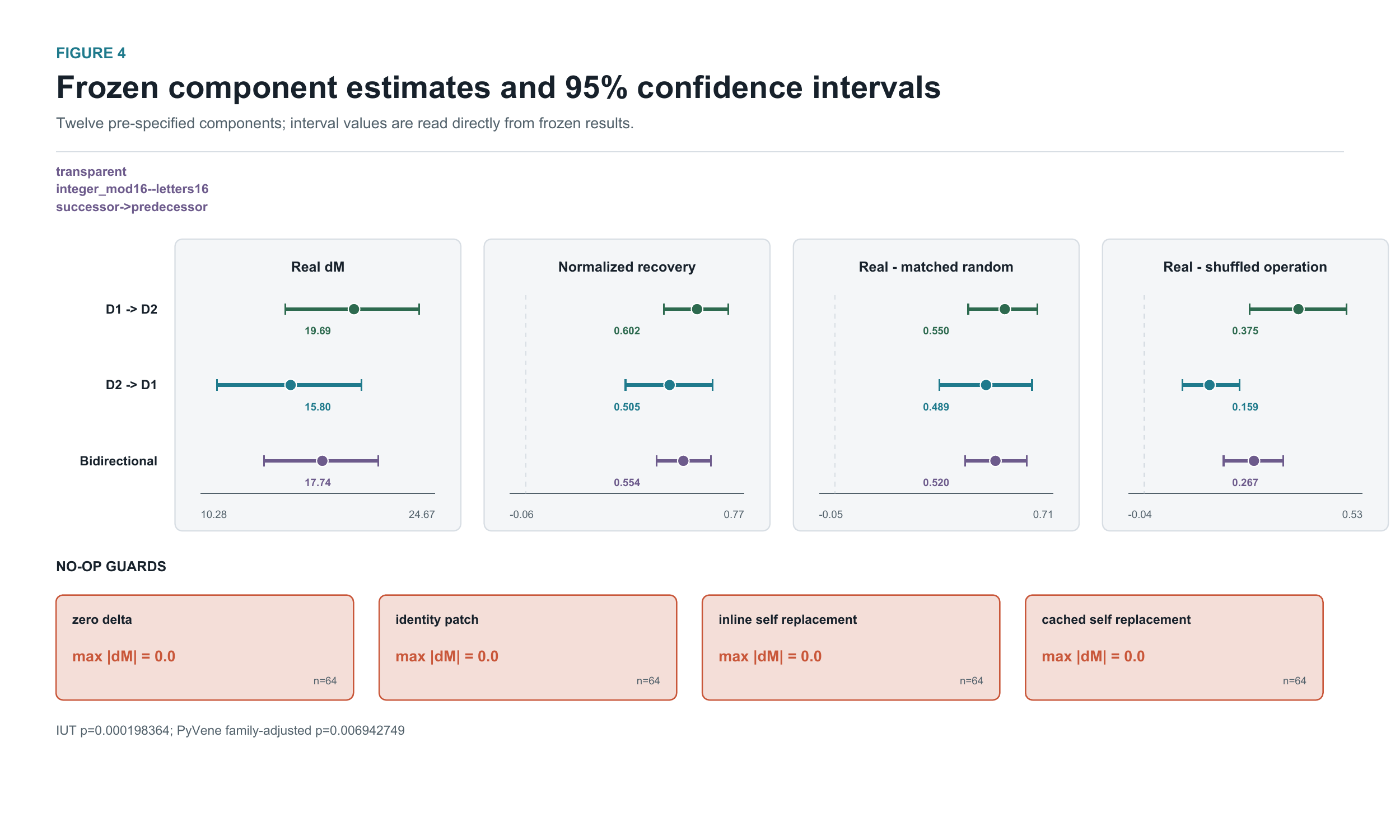}
  \caption{Frozen confirmation estimates and 95\% confidence intervals for
  all 12 pre-specified components of \candidateid. Points are frozen estimates
  and bars are frozen intervals; no bootstrap was recomputed for this figure.
  The four no-op classes have exact zero effect. The 32 matched-random
  replicates are controls, not independent samples. The NNsight selected-route
  confirmation reproduced this complete 12-component effect object
  numerically. The in-figure Holm label is the PyVene full-family value; the
  NNsight selected-route family adjustment is reported separately below.}
  \label{fig:components}
\end{figure}

\subsection{Cross-backend single-route replication with NNsight}

The NNsight implementation passed the selected route, whose criteria were
pre-specified and frozen before confirmation access, in both construction and
independently isolated confirmation. Each split contained 2,432
causal/control rows. All 12 component lower confidence bounds were positive in
construction and all 12 were positive in confirmation; the minimum lower
bounds were 0.16651101479002609 and 0.09282625375164404, respectively. No-op
failures were 0/0. Confirmation yielded
$p_{\mathrm{IUT}}=13/65536=0.0001983642578125$ and
$p_{\mathrm{Holm}}=468/65536=0.00714111328125$. The frozen
terminal status was \nnsightstatus, denoting passage of all selected-route
criteria in both NNsight splits.

For the selected route, the NNsight confirmation object matched the PyVene
confirmation object on all 12 estimates, lower and upper 95\% confidence
bounds, and exact one-sided sign-flip p-values. This equality is limited to the
pre-specified 12-component confirmation effect object: it does not assert
identity of backend-internal raw tensors or replication on independently
sampled data. The two family-adjusted p-values remain distinct for a mechanical
reason rather than a discrepancy in the effect object. In the measured PyVene
full family, the candidate's Holm rank
used multiplier 35, giving
$35\times13/65536=0.0069427490234375$. In the pre-specified NNsight
single-route family, the 35 unmeasured candidates remained at $p=1$, so the
selected candidate used multiplier 36, giving
$36\times13/65536=0.00714111328125$.

\section{Pre-Specification, Split Isolation, and Reproducibility}

The preliminary stages changed the design of the formal experiment rather than
contributing evidence to its result. They showed that semantically plausible
answers had to be separated from exact canonical outputs, that causal
measurement required state-level clean eligibility, and that a raw patching
effect needed controls for propagation, direction magnitude, operation
identity, and the patching pipeline itself. These lessons motivated the
eligibility rules, contrastive estimands, matched controls, and no-op checks used
in P3C. Detailed engineering failures and recovery procedures are retained in
the private audit history, not presented as scientific contributions.

To reduce outcome-dependent choices, the 36-candidate family, source--recipient
map, layer rule, controls, component tests, multiplicity correction, and
diagnostic boundary were specified before confirmation data became available.
These choices were recorded in a private content-addressed audit trail frozen
before confirmation access; this was not a public preregistration registry.
Construction and confirmation used separately computed eligibility and were
adjudicated independently; their pass sets were intersected only afterward.
This design provides an auditable temporal separation between hypothesis
construction and confirmation. It does not eliminate all researcher degrees of
freedom and is not a substitute for replication by an independent research
group.

A separate standard-library-only audit then recomputed the paper's numerical
claims without importing the original adjudicator or rerunning model inference.
It reconstructed the candidate family and split-specific pass sets, verified
the clean and causal/control record structure, and reproduced the 12 component
statistics, IUT and Holm values, figures, tables, and manuscript numeric traces.
This is an internal reproducibility check, not external replication.
\section{Discussion}

\subsection{What the data support}

One formal candidate from the family pre-specified and frozen before
confirmation access passed in both the construction PyVene split and the
independently isolated confirmation PyVene split. It passed under
split-specific clean eligibility, operation-specific controls, a
12-component IUT, and 36-family Holm correction.

The same frozen candidate, model revision, and layer interval subsequently
passed an NNsight selected-route construction--confirmation replication whose
scope and criteria were pre-specified and frozen before confirmation access.
This supplies replication across two intervention
implementations, but it is not a second full-family candidate screen and does
not establish backend independence.

Read narrowly, the result is consistent with a causal-transfer candidate that
is input-dependent, operation-level, and recipient-per-input. The source
contrast was extracted separately for each input state,
and the main effect was additionally constrained by matched-random,
shuffled-operation, zero/identity, and self-replacement paths. The tested
controls constrain random-direction, wrong-operation, and no-op explanations;
they do not exclude every alternative mechanism.

\subsection{Boundary relative to neighboring work}

The result should not be described as the discovery of activation steering,
successor representations, isomorphic-domain transfer, or strict controls in
themselves. Function-vector and activation-steering work already establishes
causal efficacy for internal directions; Successor Heads and
symbolic-mechanism work study ordered transformations; mOthello and
MetaOthello already establish direct precedents for causal transfer across
remapped or isomorphic domains. The contribution considered here lies only in
the joint restriction of recipient-per-input operation contrasts, frozen
operation-specific controls, isolated splits, and multiplicity-corrected
adjudication.

\subsection{Limitations}

The claim ceiling is
\emph{route-bound, single-candidate replication across two intervention
implementations, on one frozen model revision and one frozen layer interval}.
The evidence does not establish:
\begin{itemize}
  \item backend independence or full-family cross-backend replication---the
  NNsight chain was a pre-specified single-route replication, not a second
  36-candidate screen;
  \item cross-model generalization;
  \item coverage across all domains or operations;
  \item a causal effect for the diagnostic abstract route;
  \item algebraic relation preservation;
  \item a universal internal language or a ``first word.''
\end{itemize}
There is no independent identity-function control. Only one candidate
replicated across splits. Two confirmation-only positives remain frozen,
uninterpreted, and ineligible for promotion.

The direction-specific operation margin was materially asymmetric:
real-minus-shuffled recovery was 0.37498 for D1$\rightarrow$D2 but 0.15868
[0.09283, 0.23190] for D2$\rightarrow$D1. The shuffled-operation control itself
retained approximately 0.287 aggregate normalized recovery, so the evidence
does not attribute all observed recovery to an operation-specific component.

\begin{figure}[t]
  \centering
  \includegraphics[width=0.90\textwidth]{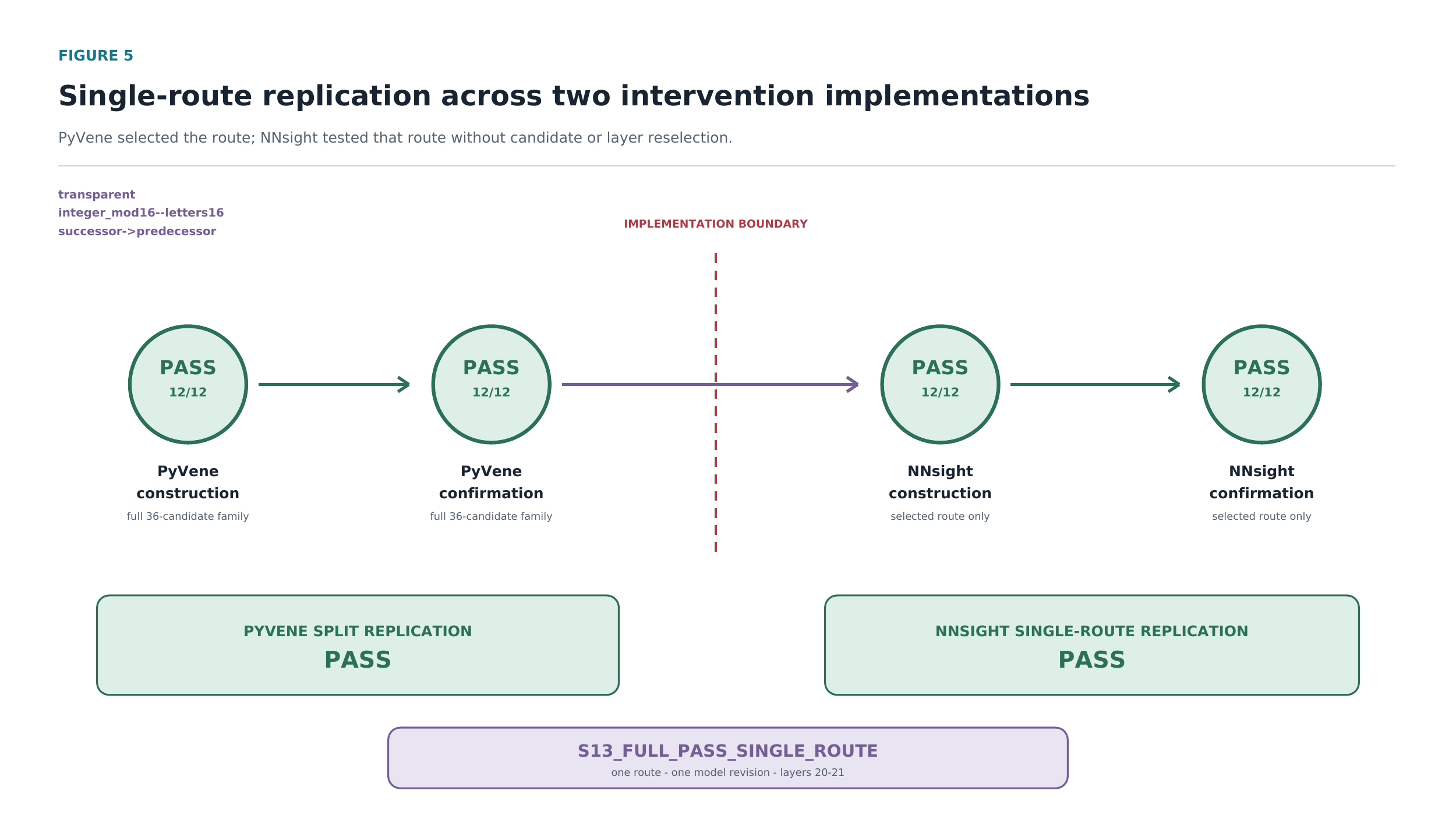}
  \caption{Current stage boundary. PyVene full-family construction and
  independently isolated confirmation produced one both-split candidate. A
  separate NNsight 0.7.0 selected-route replication, pre-specified and frozen
  before confirmation access, then passed
  construction and independently isolated confirmation for the same candidate, model
  revision, and layers. NNsight did not repeat 36-candidate screening; the 35
  unmeasured family members remained at $p=1$. The scientific terminal is
  \nnsightstatus. This does not establish backend independence or full-family
  cross-backend replication.}
  \label{fig:stage-boundary}
\end{figure}

\section{Responsible-Use Statement}

Mechanistic intervention methods can support auditing and scientific
understanding, but they can also be misused to overstate causal explanations
or steer model behavior without adequate validation. The present study
mitigates these risks through candidate and control families pre-specified and
frozen before confirmation access,
construction-confirmation isolation, explicit no-op and wrong-operation
controls, multiplicity correction, and a claim boundary that prohibits
cross-model, domain-general, full-family backend-independent, or algebraic
conclusions. The
experiments concern finite synthetic symbolic tasks and do not evaluate
deployment decisions or human-facing applications. Any future use should
revalidate effects for the target model, backend, task distribution, and
operational context.

\section{Use of AI Systems}

SHAN XINYI is the sole human author and accepts responsibility for the study and manuscript. She set the research question, scope, and decision criteria; authorized executions; approved evidence and interpretation boundaries; froze versions; reviewed verification reports; and approved the final interpretations and claims. AI systems assisted with research ideation, implementation and validation code, human-gated cloud execution, statistical pipelines, literature search, manuscript and figure drafting, engineering audit, and role-separated cross-system adversarial review. OpenAI ChatGPT, reported by the author as GPT-5.6 Sol (internal label \texttt{Chenxing (Chat)}), supported scientific synthesis, result interpretation, and open-ended critical review. OpenAI Codex, bound by the sealed project record as GPT-5.6 Sol (\texttt{Chenxing (Codex)}), supported implementation, execution control, evidence provenance, and deterministic verification. Anthropic Claude Code, bound by the sealed project record as Claude Fable 5 (\texttt{Zhenxing}), supported methodological adjudication, chain-level audit, and adversarial cold review. All registered statistics were computed by deterministic programs from sealed execution outputs and checked through separate deterministic recomputation. Numerical evidence came from execution on real hardware. Hashes establish artifact identity and non-drift, not scientific correctness. AI systems are not authors and held no autonomous authorization or publication authority. This internal cross-system review does not replace external human replication, independent data replication, or peer review.

\section*{Acknowledgment}

The author used Anthropic Claude.ai, reported by the author as Claude Opus 4.6 (internal label \texttt{Yao}), solely for non-technical reflective and motivational support. This system did not contribute to research design, implementation, data analysis, or scientific claims.

\section{Code and Data Availability}

The manuscript source, component-level tables embedded in that source,
existing ancillary CSV files, and public-safe verification materials accompany
this manuscript. The private confirmation workspace, raw tensors,
credentials, host identity, private paths, and governance records are not
public. The available materials support inspection of the reported methods,
component-level results, and deterministic numerical checks; they are not
claimed to support an independent end-to-end rerun of all experiments. No
public code repository or release date is promised here.

\section{Conclusion}

One route-bound candidate passed both construction and independently isolated
confirmation in PyVene, under a frozen 36-candidate family, independent
split-specific eligibility, operation-specific controls, and
multiplicity-corrected adjudication. A separate NNsight 0.7.0 chain,
pre-specified and frozen before confirmation access, then reproduced the same
selected route across construction and independently
isolated confirmation, including the complete 12-component confirmation effect object.
This is an existence result for one route replicated across two intervention
implementations on one model revision and layers 20--21. It is not evidence of
a universal internal language, cross-model generalization, backend
independence, full-family cross-backend replication, or preserved algebraic
structure.

\appendix
\section{Supplementary Tables}

The following tables are generated mechanically from the frozen derived-data
baseline. Full machine-readable supplements S1--S4 are provided as ancillary
files in the \texttt{anc/} directory. They are not new analyses.

\subsection{Stage roles}
\label{app:stage-roles}

\begin{table}[ht]
\centering
\small
\setlength{\tabcolsep}{3pt}
\caption{Stage roles and model separation.}
\begin{tabular}{p{0.10\textwidth}p{0.21\textwidth}p{0.46\textwidth}p{0.09\textwidth}}
\toprule
Stage & Model & Role & Causal stage entered \\
\midrule
P3A & \texttt{Qwen/\allowbreak{}Qwen2.5\allowbreak{}-\allowbreak{}1.5B-\allowbreak{}Instru\allowbreak{}ct} & 1.5B behavioral capability qualification only & no \\
P3B & \texttt{Qwen/\allowbreak{}Qwen2.5\allowbreak{}-\allowbreak{}7B-\allowbreak{}Instruct\allowbreak{}} & 7B clean screen and design narrowing; no downstream causal execution & no \\
P3C & \texttt{Qwen/\allowbreak{}Qwen2.5\allowbreak{}-\allowbreak{}7B-\allowbreak{}Instruct\allowbreak{}} & 7B formal causal construction and independent confirmation & yes \\
\bottomrule
\end{tabular}
\end{table}

\subsection{Preliminary-stage process metrics}
\label{app:preliminary-metrics}

P3A used Qwen2.5-1.5B-Instruct as a capability and measurement gate.
Semantic accuracy was 0.9175, strict generation accuracy was 0.9150, and nine
of ten task families passed. These results established task feasibility only;
they do not support the later causal result on the 7B model.

P3B then organized the three 16-state domains and three operations, fixed the
task and patching rules, and completed a 720-sample nonblind clean screen on
Qwen2.5-7B-Instruct. Canonical exact accuracy was 0.6056; 11/36 behavior cells
and 33/144 atomic-pair cells passed. Because the strict gate remained closed,
P3B did not enter its planned blinded causal stage.

\subsection{Frozen execution contract}
\label{app:execution-contract}

\begin{longtable}{p{0.29\textwidth}p{0.65\textwidth}}
\caption{Selected frozen execution identities.}\\
\toprule
Field & Value \\
\midrule
\endfirsthead
\toprule Field & Value \\ \midrule
\endhead
\texttt{model\_\allowbreak{}id} & \texttt{Qwen/\allowbreak{}Qwen2.5\allowbreak{}-\allowbreak{}7B-\allowbreak{}Instruct\allowbreak{}} \\
\texttt{model\_\allowbreak{}revisi\allowbreak{}on} & \texttt{a09a35458c70\allowbreak{}2b33eeacc393\allowbreak{}d103063234e8\allowbreak{}bc28} \\
\texttt{snapshot\_\allowbreak{}fil\allowbreak{}e\_\allowbreak{}count} & \texttt{14} \\
\texttt{snapshot\_\allowbreak{}fil\allowbreak{}e\_\allowbreak{}manifest\_\allowbreak{}s\allowbreak{}ha256} & \texttt{4ae733cf244e\allowbreak{}1ae3715c9f79\allowbreak{}a95e804d2c08\allowbreak{}fa352d5bd902\allowbreak{}30efa8f96720\allowbreak{}0336} \\
\texttt{backend} & \texttt{pyvene} \\
\texttt{backend\_\allowbreak{}vers\allowbreak{}ion} & \texttt{0.1.8} \\
\texttt{backend\_\allowbreak{}comm\allowbreak{}it} & \texttt{8138d93a7e5b\allowbreak{}1b5742d71f9f\allowbreak{}750447f2e1b6\allowbreak{}5b95} \\
\texttt{python} & \texttt{3.11.9} \\
\texttt{torch} & \texttt{2.11.0+cu128\allowbreak{}} \\
\texttt{transformers\allowbreak{}} & \texttt{4.57.6} \\
\texttt{tokenizers} & \texttt{0.22.2} \\
\texttt{gpu} & \texttt{"driver":\allowbreak{}"58\allowbreak{}0.105.08",\allowbreak{}"name":\allowbreak{}"NVIDIA GeFo\allowbreak{}rce RTX 3090\allowbreak{}"} \\
\texttt{patch\_\allowbreak{}site} & \texttt{assistant\_\allowbreak{}ou\allowbreak{}tput\_\allowbreak{}boundar\allowbreak{}y} \\
\texttt{model\_\allowbreak{}instan\allowbreak{}ce\_\allowbreak{}transform\allowbreak{}er\_\allowbreak{}block\_\allowbreak{}cou\allowbreak{}nt} & \texttt{28} \\
\texttt{selected\_\allowbreak{}lay\allowbreak{}ers\_\allowbreak{}zero\_\allowbreak{}bas\allowbreak{}ed} & \texttt{[20,\allowbreak{}21]\allowbreak{}} \\
\bottomrule
\end{longtable}

\subsection{Control roster}
\label{app:control-roster}

\begin{longtable}{p{0.21\textwidth}rrp{0.50\textwidth}}
\caption{Per-bucket and total control roster.}\\
\toprule
Type & Per bucket & Total & Generation rule \\
\midrule
\endfirsthead
\toprule Type & Per bucket & Total & Generation rule \\ \midrule
\endhead
\texttt{real} & 1 & 336 & measured source operation contrast \\
\texttt{matched\_\allowbreak{}rand\allowbreak{}om} & 32 & 10752 & deterministic seeded random directions with exact FP16 registered-norm match \\
\texttt{shuffled\_\allowbreak{}ope\allowbreak{}ration} & 1 & 336 & no-fixed-point mapping over six ordered operation transitions \\
\texttt{zero\_\allowbreak{}delta} & 1 & 336 & zero intervention payload \\
\texttt{identity\_\allowbreak{}pat\allowbreak{}ch} & 1 & 336 & identity patch \\
\texttt{inline\_\allowbreak{}self\_\allowbreak{}replacement} & 1 & 336 & inline replacement by the same activation \\
\texttt{cached\_\allowbreak{}self\_\allowbreak{}replacement} & 1 & 336 & cached replacement by the same activation \\
\bottomrule
\end{longtable}

\subsection{Confirmation record composition}
\label{app:record-composition}

\begin{table}[ht]
\centering
\small
\caption{Clean and causal/control record composition.}
\begin{tabular}{llrrrr}
\toprule
Class & Subclass & Rows & Cells/buckets & Exact & Eligible \\
\midrule
clean & \texttt{formal} & 288 & 18 & 243 & 6 \\
clean & \texttt{diagnostic} & 144 & 9 & 95 & 0 \\
causal\_control & \texttt{all} & 12768 & 336 &  &  \\
causal\_control & \texttt{cached\_\allowbreak{}self\_\allowbreak{}replacement} & 336 & 336 &  &  \\
causal\_control & \texttt{identity\_\allowbreak{}pat\allowbreak{}ch} & 336 & 336 &  &  \\
causal\_control & \texttt{inline\_\allowbreak{}self\_\allowbreak{}replacement} & 336 & 336 &  &  \\
causal\_control & \texttt{matched\_\allowbreak{}rand\allowbreak{}om} & 10752 & 336 &  &  \\
causal\_control & \texttt{real} & 336 & 336 &  &  \\
causal\_control & \texttt{shuffled\_\allowbreak{}ope\allowbreak{}ration} & 336 & 336 &  &  \\
causal\_control & \texttt{zero\_\allowbreak{}delta} & 336 & 336 &  &  \\
\bottomrule
\end{tabular}
\end{table}

\subsection{Full 36-candidate split adjudication}
\label{app:family}

\fontsize{7}{8}\selectfont
\setlength{\tabcolsep}{2pt}
\begin{longtable}{p{0.10\textwidth}p{0.16\textwidth}p{0.14\textwidth}p{0.05\textwidth}p{0.06\textwidth}p{0.06\textwidth}p{0.13\textwidth}>{\raggedleft\arraybackslash}p{0.075\textwidth}>{\raggedleft\arraybackslash}p{0.075\textwidth}}
\caption{Construction and confirmation status for all 36 formal candidates.}\\
\toprule
Route & Domains & Transition & C pass & Conf. elig. & Conf. pass & Class & IUT $p$ & Holm $p$ \\
\midrule
\endfirsthead
\toprule Route & Domains & Transition & C pass & Conf. elig. & Conf. pass & Class & IUT $p$ & Holm $p$ \\ \midrule
\endhead
\texttt{transparent} & \texttt{binary4-\allowbreak{}-\allowbreak{}int\allowbreak{}eger\_\allowbreak{}mod16} & \texttt{successor-\allowbreak{}>\allowbreak{}p\allowbreak{}redecessor} & no & no & no & \texttt{neither} & 1 & 1 \\
\texttt{transparent} & \texttt{binary4-\allowbreak{}-\allowbreak{}int\allowbreak{}eger\_\allowbreak{}mod16} & \texttt{successor-\allowbreak{}>\allowbreak{}r\allowbreak{}eflection} & no & no & no & \texttt{neither} & 1 & 1 \\
\texttt{transparent} & \texttt{binary4-\allowbreak{}-\allowbreak{}int\allowbreak{}eger\_\allowbreak{}mod16} & \texttt{predecessor-\allowbreak{}>\allowbreak{}successor} & no & no & no & \texttt{neither} & 1 & 1 \\
\texttt{transparent} & \texttt{binary4-\allowbreak{}-\allowbreak{}int\allowbreak{}eger\_\allowbreak{}mod16} & \texttt{predecessor-\allowbreak{}>\allowbreak{}reflection} & no & no & no & \texttt{neither} & 1 & 1 \\
\texttt{transparent} & \texttt{binary4-\allowbreak{}-\allowbreak{}int\allowbreak{}eger\_\allowbreak{}mod16} & \texttt{reflection-\allowbreak{}>\allowbreak{}successor} & no & no & no & \texttt{neither} & 1 & 1 \\
\texttt{transparent} & \texttt{binary4-\allowbreak{}-\allowbreak{}int\allowbreak{}eger\_\allowbreak{}mod16} & \texttt{reflection-\allowbreak{}>\allowbreak{}predecessor} & no & no & no & \texttt{neither} & 1 & 1 \\
\texttt{transparent} & \texttt{binary4-\allowbreak{}-\allowbreak{}let\allowbreak{}ters16} & \texttt{successor-\allowbreak{}>\allowbreak{}p\allowbreak{}redecessor} & no & no & no & \texttt{neither} & 1 & 1 \\
\texttt{transparent} & \texttt{binary4-\allowbreak{}-\allowbreak{}let\allowbreak{}ters16} & \texttt{successor-\allowbreak{}>\allowbreak{}r\allowbreak{}eflection} & no & no & no & \texttt{neither} & 1 & 1 \\
\texttt{transparent} & \texttt{binary4-\allowbreak{}-\allowbreak{}let\allowbreak{}ters16} & \texttt{predecessor-\allowbreak{}>\allowbreak{}successor} & no & no & no & \texttt{neither} & 1 & 1 \\
\texttt{transparent} & \texttt{binary4-\allowbreak{}-\allowbreak{}let\allowbreak{}ters16} & \texttt{predecessor-\allowbreak{}>\allowbreak{}reflection} & no & no & no & \texttt{neither} & 1 & 1 \\
\texttt{transparent} & \texttt{binary4-\allowbreak{}-\allowbreak{}let\allowbreak{}ters16} & \texttt{reflection-\allowbreak{}>\allowbreak{}successor} & no & no & no & \texttt{neither} & 1 & 1 \\
\texttt{transparent} & \texttt{binary4-\allowbreak{}-\allowbreak{}let\allowbreak{}ters16} & \texttt{reflection-\allowbreak{}>\allowbreak{}predecessor} & no & no & no & \texttt{neither} & 1 & 1 \\
\texttt{transparent} & \texttt{integer\_\allowbreak{}mod1\allowbreak{}6-\allowbreak{}-\allowbreak{}letters16\allowbreak{}} & \texttt{successor-\allowbreak{}>\allowbreak{}p\allowbreak{}redecessor} & yes & yes & yes & \texttt{both\_\allowbreak{}split\_\allowbreak{}p\allowbreak{}ass} & 0.000198 & 0.006943 \\
\texttt{transparent} & \texttt{integer\_\allowbreak{}mod1\allowbreak{}6-\allowbreak{}-\allowbreak{}letters16\allowbreak{}} & \texttt{successor-\allowbreak{}>\allowbreak{}r\allowbreak{}eflection} & yes & no & no & \texttt{construction\allowbreak{}\_\allowbreak{}only} & 1 & 1 \\
\texttt{transparent} & \texttt{integer\_\allowbreak{}mod1\allowbreak{}6-\allowbreak{}-\allowbreak{}letters16\allowbreak{}} & \texttt{predecessor-\allowbreak{}>\allowbreak{}successor} & yes & yes & no & \texttt{construction\allowbreak{}\_\allowbreak{}only} & 0.00293 & 0.09668 \\
\texttt{transparent} & \texttt{integer\_\allowbreak{}mod1\allowbreak{}6-\allowbreak{}-\allowbreak{}letters16\allowbreak{}} & \texttt{predecessor-\allowbreak{}>\allowbreak{}reflection} & no & no & no & \texttt{neither} & 1 & 1 \\
\texttt{transparent} & \texttt{integer\_\allowbreak{}mod1\allowbreak{}6-\allowbreak{}-\allowbreak{}letters16\allowbreak{}} & \texttt{reflection-\allowbreak{}>\allowbreak{}successor} & yes & no & no & \texttt{construction\allowbreak{}\_\allowbreak{}only} & 1 & 1 \\
\texttt{transparent} & \texttt{integer\_\allowbreak{}mod1\allowbreak{}6-\allowbreak{}-\allowbreak{}letters16\allowbreak{}} & \texttt{reflection-\allowbreak{}>\allowbreak{}predecessor} & no & no & no & \texttt{neither} & 1 & 1 \\
\texttt{relational} & \texttt{binary4-\allowbreak{}-\allowbreak{}int\allowbreak{}eger\_\allowbreak{}mod16} & \texttt{successor-\allowbreak{}>\allowbreak{}p\allowbreak{}redecessor} & no & no & no & \texttt{neither} & 1 & 1 \\
\texttt{relational} & \texttt{binary4-\allowbreak{}-\allowbreak{}int\allowbreak{}eger\_\allowbreak{}mod16} & \texttt{successor-\allowbreak{}>\allowbreak{}r\allowbreak{}eflection} & no & yes & yes & \texttt{confirmation\allowbreak{}\_\allowbreak{}only\_\allowbreak{}frozen\allowbreak{}} & 0.000488 & 0.016602 \\
\texttt{relational} & \texttt{binary4-\allowbreak{}-\allowbreak{}int\allowbreak{}eger\_\allowbreak{}mod16} & \texttt{predecessor-\allowbreak{}>\allowbreak{}successor} & no & no & no & \texttt{neither} & 1 & 1 \\
\texttt{relational} & \texttt{binary4-\allowbreak{}-\allowbreak{}int\allowbreak{}eger\_\allowbreak{}mod16} & \texttt{predecessor-\allowbreak{}>\allowbreak{}reflection} & no & no & no & \texttt{neither} & 1 & 1 \\
\texttt{relational} & \texttt{binary4-\allowbreak{}-\allowbreak{}int\allowbreak{}eger\_\allowbreak{}mod16} & \texttt{reflection-\allowbreak{}>\allowbreak{}successor} & yes & yes & no & \texttt{construction\allowbreak{}\_\allowbreak{}only} & 0.010742 & 0.34375 \\
\texttt{relational} & \texttt{binary4-\allowbreak{}-\allowbreak{}int\allowbreak{}eger\_\allowbreak{}mod16} & \texttt{reflection-\allowbreak{}>\allowbreak{}predecessor} & no & no & no & \texttt{neither} & 1 & 1 \\
\texttt{relational} & \texttt{binary4-\allowbreak{}-\allowbreak{}let\allowbreak{}ters16} & \texttt{successor-\allowbreak{}>\allowbreak{}p\allowbreak{}redecessor} & no & no & no & \texttt{neither} & 1 & 1 \\
\texttt{relational} & \texttt{binary4-\allowbreak{}-\allowbreak{}let\allowbreak{}ters16} & \texttt{successor-\allowbreak{}>\allowbreak{}r\allowbreak{}eflection} & no & no & no & \texttt{neither} & 1 & 1 \\
\texttt{relational} & \texttt{binary4-\allowbreak{}-\allowbreak{}let\allowbreak{}ters16} & \texttt{predecessor-\allowbreak{}>\allowbreak{}successor} & no & no & no & \texttt{neither} & 1 & 1 \\
\texttt{relational} & \texttt{binary4-\allowbreak{}-\allowbreak{}let\allowbreak{}ters16} & \texttt{predecessor-\allowbreak{}>\allowbreak{}reflection} & no & no & no & \texttt{neither} & 1 & 1 \\
\texttt{relational} & \texttt{binary4-\allowbreak{}-\allowbreak{}let\allowbreak{}ters16} & \texttt{reflection-\allowbreak{}>\allowbreak{}successor} & no & no & no & \texttt{neither} & 1 & 1 \\
\texttt{relational} & \texttt{binary4-\allowbreak{}-\allowbreak{}let\allowbreak{}ters16} & \texttt{reflection-\allowbreak{}>\allowbreak{}predecessor} & no & no & no & \texttt{neither} & 1 & 1 \\
\texttt{relational} & \texttt{integer\_\allowbreak{}mod1\allowbreak{}6-\allowbreak{}-\allowbreak{}letters16\allowbreak{}} & \texttt{successor-\allowbreak{}>\allowbreak{}p\allowbreak{}redecessor} & no & yes & yes & \texttt{confirmation\allowbreak{}\_\allowbreak{}only\_\allowbreak{}frozen\allowbreak{}} & 0.000183 & 0.006592 \\
\texttt{relational} & \texttt{integer\_\allowbreak{}mod1\allowbreak{}6-\allowbreak{}-\allowbreak{}letters16\allowbreak{}} & \texttt{successor-\allowbreak{}>\allowbreak{}r\allowbreak{}eflection} & no & no & no & \texttt{neither} & 1 & 1 \\
\texttt{relational} & \texttt{integer\_\allowbreak{}mod1\allowbreak{}6-\allowbreak{}-\allowbreak{}letters16\allowbreak{}} & \texttt{predecessor-\allowbreak{}>\allowbreak{}successor} & no & yes & no & \texttt{neither} & 0.06781 & 1 \\
\texttt{relational} & \texttt{integer\_\allowbreak{}mod1\allowbreak{}6-\allowbreak{}-\allowbreak{}letters16\allowbreak{}} & \texttt{predecessor-\allowbreak{}>\allowbreak{}reflection} & no & no & no & \texttt{neither} & 1 & 1 \\
\texttt{relational} & \texttt{integer\_\allowbreak{}mod1\allowbreak{}6-\allowbreak{}-\allowbreak{}letters16\allowbreak{}} & \texttt{reflection-\allowbreak{}>\allowbreak{}successor} & no & no & no & \texttt{neither} & 1 & 1 \\
\texttt{relational} & \texttt{integer\_\allowbreak{}mod1\allowbreak{}6-\allowbreak{}-\allowbreak{}letters16\allowbreak{}} & \texttt{reflection-\allowbreak{}>\allowbreak{}predecessor} & no & no & no & \texttt{neither} & 1 & 1 \\
\bottomrule
\end{longtable}
\normalsize

\subsection{Twelve pre-specified components}
\label{app:components}

\small
\setlength{\tabcolsep}{3pt}
\begin{longtable}{p{0.14\textwidth}p{0.30\textwidth}>{\raggedleft\arraybackslash}p{0.10\textwidth}>{\raggedleft\arraybackslash}p{0.10\textwidth}>{\raggedleft\arraybackslash}p{0.10\textwidth}>{\raggedleft\arraybackslash}p{0.13\textwidth}}
\caption{Frozen component estimates, 95\% intervals, and exact one-sided sign-flip p-values.}\\
\toprule
Scope & Metric & Estimate & Lower & Upper & Exact $p$ \\
\midrule
\endfirsthead
\toprule Scope & Metric & Estimate & Lower & Upper & Exact $p$ \\ \midrule
\endhead
D1 to D2 & real dM & 19.690188 & 15.485354 & 23.678714 & 1.5258789e-05 \\
D1 to D2 & normalized recovery & 0.601868 & 0.484423 & 0.711212 & 1.5258789e-05 \\
D1 to D2 & real - mean matched random & 0.549932 & 0.431464 & 0.655376 & 1.5258789e-05 \\
D1 to D2 & real - shuffled operation & 0.374984 & 0.256848 & 0.492048 & 6.1035156e-05 \\
D2 to D1 & real dM & 15.799307 & 11.270405 & 20.154126 & 1.5258789e-05 \\
D2 to D1 & normalized recovery & 0.505288 & 0.350213 & 0.656517 & 1.5258789e-05 \\
D2 to D1 & real - mean matched random & 0.489285 & 0.337995 & 0.638441 & 1.5258789e-05 \\
D2 to D1 & real - shuffled operation & 0.15868 & 0.092826 & 0.231904 & 0.000198 \\
bidirectional aggregate & real dM & 17.744748 & 14.174127 & 21.172185 & 1.5258789e-05 \\
bidirectional aggregate & normalized recovery & 0.553578 & 0.45874 & 0.650427 & 1.5258789e-05 \\
bidirectional aggregate & real - mean matched random & 0.519609 & 0.421521 & 0.620572 & 1.5258789e-05 \\
bidirectional aggregate & real - shuffled operation & 0.266832 & 0.192949 & 0.338611 & 1.5258789e-05 \\
\bottomrule
\end{longtable}
\normalsize

\subsection{Claim boundary}
\label{app:claim-boundary}

\small
\begin{longtable}{p{0.22\textwidth}p{0.34\textwidth}p{0.36\textwidth}}
\caption{Supported claim and prohibited extrapolations.}\\
\toprule
Class & Evidence status & Permitted or prohibited wording \\
\midrule
\endfirsthead
\toprule Class & Evidence status & Permitted or prohibited wording \\ \midrule
\endhead
supported & transparent|integer\_mod16--letters16|successor->predecessor & route-bound, single-candidate replication across two intervention implementations, on one frozen model revision and one frozen layer interval \\
protocol terminal success & MET WITHIN FROZEN S13 SINGLE-ROUTE SCOPE --- S13\_\allowbreak{}FULL\_\allowbreak{}PASS\_\allowbreak{}SINGLE\_\allowbreak{}ROUTE & May describe terminal success only for the frozen selected route and claim boundary. \\
cross backend replication & SUPPORTED ONLY AS SINGLE-ROUTE REPLICATION ACROSS TWO INTERVENTION IMPLEMENTATIONS --- NNsight did not repeat full 36-candidate screening & May claim selected-route replication across PyVene and NNsight implementations; do not claim backend independence or full-family cross-backend replication. \\
cross model generalization & NOT TESTED --- formal P3C is bound to the frozen Qwen2.5-7B-Instruct revision & Do not claim cross-model generalization. \\
all domains all operations & NOT SUPPORTED --- only one complete candidate is split-replicated & Do not claim support across all domains or all operations. \\
abstract route causal confirmation & NOT SUPPORTED --- the diagnostic route is non-promotional and had zero eligible candidates & Do not claim abstract-route causal confirmation. \\
algebraic relation preservation & NOT TESTED --- composition, inverse, involution, and conjugacy relations remain untested & Do not claim algebraic relation preservation. \\
universal language or first word & PROHIBITED SCIENTIFIC CLAIM --- internal metaphor only; not a scientific result & Do not use 'universal language' or 'first word' as a scientific claim. \\
confirmation only positives & FROZEN / UNINTERPRETED / NO PROMOTION --- relational|binary4--integer\_mod16|successor->reflection; relational|integer\_mod16--letters16|successor->predecessor & Do not interpret or promote confirmation-only positives into the current claim. \\
\bottomrule
\end{longtable}
\normalsize

\subsection{Registered FP16 norm contract}
\label{app:fp16}

\begin{table}[ht]
\centering
\small
\caption{Final registered-norm contract. Retry-development history is not part of the scientific claim.}
\begin{tabular}{ll}
\toprule
Field & Value \\
\midrule
\texttt{comparison} & \texttt{exact int16 \allowbreak{}bit-\allowbreak{}pattern \allowbreak{}equality} \\
\texttt{control\_\allowbreak{}vect\allowbreak{}or\_\allowbreak{}persisted\allowbreak{}\_\allowbreak{}on\_\allowbreak{}failure} & \texttt{false} \\
\texttt{failure\_\allowbreak{}sign\allowbreak{}ature} & \texttt{fp16\_\allowbreak{}registe\allowbreak{}red\_\allowbreak{}norm\_\allowbreak{}mat\allowbreak{}ch\_\allowbreak{}failed} \\
\texttt{precise\_\allowbreak{}norm\allowbreak{}\_\allowbreak{}accumulatio\allowbreak{}n\_\allowbreak{}dtype} & \texttt{float32} \\
\texttt{registered\_\allowbreak{}s\allowbreak{}calar\_\allowbreak{}dtype} & \texttt{float16} \\
\texttt{scientific\_\allowbreak{}e\allowbreak{}ffect\_\allowbreak{}value\_\allowbreak{}persisted\_\allowbreak{}on\allowbreak{}\_\allowbreak{}failure} & \texttt{false} \\
\bottomrule
\end{tabular}
\end{table}

\bibliographystyle{plainnat}
\bibliography{references}

\end{document}